\documentclass{article}

\usepackage[final]{neurips_2022}

\usepackage[utf8]{inputenc} 
\usepackage[T1]{fontenc}    
\usepackage{hyperref}       
\usepackage{url}            
\usepackage{booktabs}       
\usepackage{amsfonts}       
\usepackage{nicefrac}       
\usepackage{microtype}      
\usepackage{xcolor}         
\usepackage{multirow}
\usepackage{amsmath}
\usepackage{amsthm}
\usepackage{bbm}
\usepackage{graphicx}
\usepackage[linesnumbered,ruled,vlined]{algorithm2e}

\newtheorem{definition}{Definition}[section]

\title{Bias Mitigation Framework for Intersectional Subgroups in Neural Networks }

\author{%
  Narine Kokhlikyan \\
  Meta AI \\
  \texttt{narine@meta.com} \\
   \And
  Bilal Alsallakh \\
  Meta AI \\
  \texttt{bilalsal@meta.com} \\
  \And
  Fulton Wang \\
  Meta AI \\
  \texttt{fultonwang@meta.com} \\
  \AND
  Vivek Miglani \\
  Meta AI \\
  \texttt{vivekm@meta.com} \\
  \And
  Oliver Aobo Yang \\
  Meta AI \\
  \texttt{aoboyang@meta.com} \\
  \And
  David Adkins \\
  Meta AI \\
  \texttt{davidadkins@meta.com} \\
}

\begin{document}
\maketitle

\begin{abstract}
We propose a fairness-aware learning framework that mitigates intersectional subgroup bias associated with protected attributes.
Prior research has primarily focused on mitigating one kind of bias by incorporating complex fairness-driven constraints into optimization objectives or designing additional layers that focus on specific protected attributes. We introduce a simple and generic bias mitigation approach that prevents models from learning relationships between protected attributes and output variable by reducing mutual information between them. We demonstrate that our approach is effective in reducing bias with little or no drop in accuracy.
We also show that the models trained with our learning framework become causally fair and insensitive to the values of protected attributes.
Finally, we validate our approach by studying feature interactions between protected and non-protected attributes. We demonstrate that these interactions are significantly reduced when applying our bias mitigation.

\end{abstract}

\section{Introduction}
The unprecedented adoption of Machine Learning (ML) in critical sectors such as finance, healthcare and education has made fairness-related bias detection and mitigation a crucial part of ML systems. It is important that ML models do not discriminate against individuals based on protected attributes such as race, gender or skin color when making predictions.
Fairness research has typically focused on detecting and mitigating bias for a single protected attribute. With such detection and mitigation, the fairness gap for \emph{intersectional subgroups} such as \texttt{Female} \underline{and} \texttt{Black} might still be high. This phenomenon is called Fairness Gerrymandering~\citep{ghosh2021characterizing, pmlr-v80-kearns18a, pmlr-v81-buolamwini18a}. Several works~\citep{pmlr-v80-kearns18a, overlapping_fairness, MultiFair, fairness_auditing} proposed techniques to address fairness gerrymandering, which require predetermining fairness violation metrics. Nevertheless, a recent survey~\citep{comp_persp} highlights the scarcity of research in detecting and mitigating intersectional bias.
One of the key challenges of this research is the limited access to protected attributes and their \emph{unknown correlations} with other attributes in the dataset. These limitations are common due to privacy or data restrictions.
Another key challenge of fairness research is reducing the bias across groups and \emph{intersectional subgroups} without a significant drop in accuracy. 

In this paper we propose a generic and simple fairness-aware learning framework that addresses the intersectional subgroup bias problem without requiring specific fairness metrics to be predetermined. It learns latent representations without relying on protected attributes and their interactions with other attributes in the dataset. 
Previous work~\citep{fairness_mi, song2019learning} shows that reducing the mutual information between a protected attribute and output variable plays a significant role in reducing the bias associated with that protected attribute. We propose a generic framework that reduces the mutual information not only between a single protected attribute and the output variable, but also between any subset of protected attributes and the output variable.  Experimentally, we show that this approach debiases the model significantly with little or no drop in accuracy.
Furthermore, we also show that the models trained with our framework become insensitive to the values of protected attributes. This shows that our models are causally fair according to the definitions of causal fairness by~\citep{causal_fair}.


\paragraph{Motivating example:}
\vspace{-7pt}
Table~\ref{tab:TPRs_Male_Female} illustrates accuracy and TPR gap between two subpopulations, $\texttt{Male} \cap \texttt{White}$  and $\texttt{Female} \cap  \texttt{Black}$ for the Law School Admissions Council (LSAC) dataset~\citep{wightman1998lsac}. TPR is an important metric in this specific example since it measures the advantaged outcome.
We compare our fairness-aware framework with three other baseline models: 1) the original model without mitigation, 2) a baseline model which is trained without \texttt{Gender} and \texttt{Race} attributes and 3) a baseline model which is trained after upsampling minority subgroups.
We observe a significant TPR gap between two populations in the original model. Two of the baseline models reduce the fairness gap. However, they are less effective compared to our approach. Our approach learns latent representations that do not rely on protected attributes and their relationship to output variable, and thus reduces the fairness gap significantly.
\begin{table}
  \caption{Accuracy and True Positive Rate (TPR) for two subpopulations, computed for the original model, our model trained using fairness-aware framework and two baseline models. 
  }
  \label{tab:TPRs_Male_Female}
  \centering
  \def\arraystretch{1.0}
  \begin{tabular}{cccc}
    \toprule
    & \multirow{2}{*}{Accuracy} & \multicolumn{2}{c}{TPR} \\ 
    \cmidrule{3-4} 
     &  &  $\texttt{Male} \cap \texttt{White}$  & $\texttt{Female} \cap \texttt{Black}$ \\
    \midrule
    Original model & 0.87 & 0.96 & 0.77\\
    Removed protected attributes & 0.86 & 0.96 & 0.83 \\
    Upsampled minority groups & 0.86 & 0.88 & 0.82\\
    Our fairness-aware framework & 0.86 & 0.94 & 0.90\\
  \bottomrule
\end{tabular}
\end{table}
The main contributions of this paper are as follows:
\begin{itemize}
\vspace{-2pt}
    \item We introduce a novel bias mitigation framework that aims to reduce the mutual information between intersectional subgroups and the output variable.
    \item We show empirically that our approach reduces the equality and demographic parity gaps significantly and surpasses state-of-the-art approaches.
    \item We study the sensitivity of the debiased model to the presence of protected attributes and the effectiveness of our approach when non-protected attributes are correlated with the protected ones.
    \item We demonstrate that feature interactions between protected and non-protected attributes reduce significantly when the models are trained using our bias mitigation framework.
\end{itemize}

\section{Related Work}
Fairness literature offers numerous definitions of fairness~\citep{21_def_fairness}, its measurement and mitigation. We base our fairness definition and measurement on the work of ~\citep{Hardt2016EqualityOO} and three well-known metrics: Demographic Parity~\citep{dp, dp2, dp3}, Equalized Odds and Equality of Opportunity~\citep{Hardt2016EqualityOO}. Demographic Parity compares the average prediction score across different subgroups.
Equality of Opportunity, in addition to that, takes the label distribution into account and measures the TPR gap among different groups.
Equalized Odds~\citep{Hardt2016EqualityOO} measures both the TPR and False Positive Rate (FPR) gaps among different groups.
Specific metrics have been developed for intersectional subgroups such as the min-max ratio~\citep{ghosh2021characterizing} and differential fairness metric~\citep{intersectional_def_fairness}. In this paper we focus on measuring the commonly used Demographic Parity and Equalized Odds for intersectional subgroups, which facilitate comparisons with previous work.


Bias mitigation techniques reduce the disparities among the groups and intersectional subgroups measured by the aforementioned metrics. 
Three types of mitigation techniques have been proposed to combat fairness bias~\citep{comp_persp}: 

    \textbf{Postprocessing techniques} 
    aim to reduce fairness bias during model inference. Those approaches enforce model predictions to follow the same distribution observed during training~\citep{zhao-etal-2017-men}, transforming model predictions to follow a specific fairness measure.
    These techniques, however, require access to protected attributes during inference, which is not always available due to data scarcity or privacy reasons.

    \textbf{Dataset preprocessing techniques} such as balancing the distribution of data labels, downsampling and sample re-weighing~\citep{Kamiran2011DataPT} alleviate modelling bias to a certain extent. 
    However, \cite{balance_not_enough} shows that data preprocessing and balancing datasets often have limited effect, compared with training inherently unbiased models. Apart from data balancing, one can also delete protected attributes from the training set or mask them. However, this is not sufficient, since protected attributes are often correlated with other attributes in the data. 
    
    \textbf{Train-time techniques} aim to combat fairness bias during model training. This can be accomplished using constraints based on adversarial loss~\citep{data_decisions, Zhang2018MitigatingUB, balance_not_enough}, feature importance~\citep{Liu2019IncorporatingPW, Du2019LearningCD, right_for_right_reasons}, fairness measurement~\citep{agarwal2018reductions}, decision boundary~\citep{fairness_constraints} or statistical dependence~\citep{prejudice}. Adversarial loss requires defining additional heads or constraints for a specific protected attribute. It maximizes the primary objective of a specific task while minimizing the model's ability to predict specific protected attributes. Constraints based on feature importance, on the other hand, heavily rely on the feature contribution score
    ~\citep{Liu2019IncorporatingPW, Du2019LearningCD}, which is not always reliable~\citep{Hooker2019ABF}.
    ~\cite{overlapping_fairness} propose Bayes-optimal classification framework for intersectional group fairness which is also tied to fairness metric constraints.
    ~\cite{MultiFair} propose a framework based on mutual information minimization  for intersectional subgroups. This framework, however, requires two estimators and two additional predictors which makes the mitigation process complex.

In contrast, we propose a simple and generic training-time fairness-aware framework that doesn't rely on specific fairness metrics, or architectural modifications such as adversarial heads. It accounts for intersectional fairness of any subsets of input features and is straightforward to implement.

\vspace{-1pt}
\section{Preliminaries}
In this section we formalize the problem, introduce preliminary notations and concepts that are used later in the paper.
The goal is to learn a fair Neural Network (NN) model that mitigates the fairness gap for intersectional subgroups formed by multiple protected attributes. We seek to reach this goal with minimal impact on accuracy.
During inference, no information about the protected attributes is necessary. 

\vspace{-1pt}
\subsection{Notation}
\label{notation}
We consider a typical ML model, $f:R^M \rightarrow R^C$, that is trained on a dataset $D = \{(x_1, y_1), ..., (x_N, y_N)\}$, where each sample $x_i$ consists of a set of $M$ features $x_i = \{x_i^{1}, x_i^{2}..., x_i^{M}\}$ where $x_i^j \in R$ represents the $j^\texttt{th}$ feature in the $i^\texttt{th}$ sample. $y_i \in [1,...,C]$ is the label corresponding to sample $x_i$. Let $x'^j \in R$ denote a shared baseline across all samples for input feature $j$. It indicates the absence of signal or feature value in the input. Traditionally, the zero value is used to indicate the absence of signal but for certain features zero value might represent a meaning. For example, $0$ and $1$ might indicate male and female for the gender feature in some datasets.

\paragraph{Protected Attribute Notation.} \vspace{-7pt} Let $A_k \subseteq \{1, 2,..., M\}$ be a subset of features that are known to be protected. For example, $A_k$ may correspond to $\{race\}$ or $\{gender, race\}$. Let also $\mathcal{A}$ denote a set of subsets of $A_k$. For example, $\mathcal{A}$ may correspond to $\mathcal{A}=\{\{gender\}, \{race\}, \{gender, race\}\}$.

Finally, let also $\mathcal{S}(x_i,x',A_k)$ denote a substitution function that replaces the features that aren't in $A_k$ with values from baseline $x'$. 
\begin{equation}
\label{fig:mask_context}
    \mathcal{S}(x_i,x',A_k) = \left \{ \begin{array}{rcl}
    x_i^j, & j \in A_k \\
    x'^j, & \text{otherwise} \\
\end{array}\right. \forall j\in [M]
\end{equation}
We denote by $x^{A_k}$ the subset of features in $x$ corresponding to protected features $A_k$. Analogously, $x^{\setminus A_k}$ denotes a subset of features that excludes protected attributes $A_k$.

\vspace{-3pt}
\subsection{Information Theory}
\label{mutual_information}
We briefly revisit information theory in order to analyse the relationship between a subset of protected attributes $A_k$ and dependant output variable $y$. This analysis explains the reasoning behind our bias mitigation approach.
The entropy $H(X)$ measures the average uncertainty~\citep{10.5555/1146355} of a random variable $X$. 
Mutual Information (MI) uses the entropy to measure the shared information between two random variables. In our case, the mutual information between an input feature $x^j$ and the output variable $y$ can be measured as $MI(x^j;y) = H(x^j) - H(x^j|y)$. 
During bias mitigation, we aim to reduce the $MI$ between the protected attributes $x^{A_k}$ and $y$. Reducing $MI(x^{A_k};y)$ implies increasing the uncertainty $H(x^{A_k}|y)$. We increase that uncertainty by associating protected attributes with a uniformly distributed random output variable $Unif(1,C)$.
We propose a proxy metric that associates $x^{A_k}$ with random labels independent of the values in $x^{\setminus A_k}$. The new examples generated by $\mathcal{S}(x_i,x',A_k)$ reduce overall mutual information between $x^{A_k}$ and the protected attributes in the dataset. 
The proxy measure aims to reduce the distance between $\mathcal{S}(x_i,x',A_k)$ and $Unif(1,C)$ which can be represented as a regularization term in the optimization objective.
Table~\ref{tab:mut_info} shows how the mutual information between protected attributes $\texttt{Gender}$ and $\texttt{Race}$ reduces as we augment the data with samples that associate intersectional subgroups of protected attributes with random guesses.
In the next section we describe how we incorporate this regularization term into the model's optimization objective.

\vspace{-5pt}
\begin{table}[hbt!]
  \def\arraystretch{1.0}
  \caption{MI between protected attributes $\texttt{Gender}$ and $\texttt{Race}$ and output variable $\texttt{Passed Bar}$ before and after MI constraint-based data augmentation for the LSAC and Adult datasets.}
  \label{tab:mut_info}
  \centering
  \begin{tabular}{lcccc}
    \toprule
    & \multicolumn{2}{c}{Passed Bar (Before Mitigation)} & \multicolumn{2}{c}{Passed Bar (After Mitigation)}\\ 
     \cmidrule{2-5} 
    & LSAC & Adult & LSAC & Adult \\
    \midrule
    Gender & 0.00101 & 0.043532 & 0.00029 & 0.011177 \\
    Race & 0.02265 & 0.010469 & 0.004876 & 0.0010130 \\
  \bottomrule
\end{tabular}
\end{table}
\vspace{-7pt}
\section{Fairness-Aware Learning Algorithm}
\label{sec: fairness-learning-algo}
The learning algorithm we propose is similar in spirit to the ones which incorporate predetermined constraints into optimization objective as regularization terms.
It is more generic in its nature and does not use EO, DP or adversarial heads as a proxy for the minimization of the mutual information~\citep{fairness_mi, song2019learning} between the protected attributes and the output variable.
We give a formal definition of the objective function starting from the definition of the proxy constraint for mutual information.

\begin{definition} Given a subset of protected attributes $A_k \in \mathcal{A}$, an input example $x_i$, a uniformly random chosen label $y_{rand}$ and a baseline $x'$, the proxy constraint for the mutual information is defined as follow:
\end{definition}
\vspace{-9pt}
\begin{equation}
\label{L_A}
L^\mathcal{A}(x_i, x', \mathcal{A}, y_{rand}) = -\sum_{A_k\in \mathcal{A}} \sum_{c\in C} \mathbbm{1} (y_{rand} = c) \cdot \log(f_c(\mathcal{S}(x_i,x',A_k)))
\end{equation}
\vspace{-1pt}
where $c \in C$ are possible prediction classes and $y_{rand} = Unif(1,C)$ is a label drawn uniformly random from $C$ and $f_c$ is the NN's output for class $c$. The first summation over the subsets of protected attributes $A_k$ helps us to 
address bias mitigation for multiple subsets of protected attributes. The joint objective of a multiclass classification problem is the following.
\begin{equation}
\label{optim_loss}
L_{combined} = \sum_{(x_i, y_i) \in D} L(x_i,y_i) + \alpha \cdot \sum_{(x_i, x') \in D'} L^A(x_i, x', \mathcal{A}, Unif(1,C))
\end{equation}
\vspace{-1pt}
In our setup we use a classification loss but other loss definitions can be used instead. $L(x_i, y_i)$ represents the loss for the original model.
The hyperparameter
$\alpha$ is used to balance the amount of regularization that we incorporate into the loss. 
$D'=\{(x_1, x'), ..., (x_N, x')\}$ represents the dataset with a set of baseline values $x'=\{x'^j\}_{j=1}^{j=M}$ for each feature $j$. $Unif(1, C)$ chooses a label uniformly at random from $[1,...,C]$.~\cite{sturmfels2020visualizing, baselines} discuss different strategies for baseline selection. In a general case we can also choose different baselines for a feature and aggregate their impact on the overall loss. 
Algorithm~\ref{alg:listing} illustrates an example of how our proposed loss can be computed during training using Gradient Descent (GD) algorithm. Our method is not limited to GD and can be trained with other optimization algorithms as well. The proxy constraint plays a role of weighted data augmentation that helps the model to learn associations between protected attributes and randomly guessed labels.
\vspace{-8pt}
\SetKwInput{KwInput}{Input}                
\SetKwInput{KwOutput}{Output}              
\begin{algorithm}
\caption{Fairness-Aware Learning Algorithm}
\label{alg:listing}
\KwInput {Training Datasets $D=\{(x_i, y_i)\}_{i=1}^{i=N}$, 
Validation dataset $D^{\text{valid}}=\{(x_i, y_i)\}_{i=1}^{i=F}$, Baseline $x'=\{x'^j\}_{j=1}^{j=M}$, A set of subsets of protected attributes $\mathcal{A}$,
hyperparameter $\alpha$, learning rate $\eta$, $max\_epochs$.\\
}
\KwOutput {$W_{\text{best}}$ for the best accuracy of the model $f$ on $D^{\text{valid}}$ 
}
Initialize the model parameters $W_0$, set epoch=0; \\

\While{$epoch < max\_epochs$}
{
    $L_{\text{initial}}$ = $\frac{1}{N} \cdot \sum_{i=1}^{i=N} \sum_{c=1}^{C} \mathbbm{-1} (y_i = c) \cdot log(f_c(x_i))$;  \\
    $y_{\text{rand}} = uniform\_rand(C)$ \tcp*[r]{Uniformly at random chooses class in $[1, C]$ for each example} 
     $L^\mathcal{A}$ = $\frac{1}{N}\sum_{i=1}^{i=N} \sum_{k=1}^{|\mathcal{A}|} \sum_{c=1}^{C} \mathbbm{-1} (y_i = c) \cdot log(f_c(\mathcal{S}(x_i, x', A_k)))$ ; \\
    $L_{\text{combined}} = L_{\text{initial}} + \alpha \cdot L^\mathcal{A}$  \\
    $\text{epoch} = \text{epoch}+1$ ; \\
   $W_{\text{epoch}} = Optimizer(L_{\text{combined}}, \eta)$\tcp*[r]{Optimizer can be Adam, for example} 
    Update $W_{\text{best}}$ based on the highest accuracy measures on $D^{\text{valid}}$
    so far.
}
\end{algorithm}
\vspace{-7pt}
\section{Experiments}
In this section we present experimental results of our fairness-aware learning framework and a number of state-of-the-art approaches. 
We also describe experimental setup and discuss empirical results.
\vspace{-5pt}
\subsection{Experimental Setup}
\label{sec:setup}
The UCI Adult dataset~\citep{Dua:2019} and the Law School Admissions Council (LSAC) dataset~\citep{wightman1998lsac} are two highly unbalanced datasets used in our experiments. Appendix~\ref{dataset_details} provides additional details about those datasets. For these two datasets we built a 2-layer RELU-BatchNorm-Linear NN models similar to the one demonstrated in the literature~\citep{data_decisions}. The first linear layer contains 128, the second 64 and the last output layer only a single neuron. The models are trained using approximately 100 epochs with Adam optimizer, $0.001$ learning rate and binary cross entropy logit loss as the baseline model's loss, $L(x_i, y_i)$. As bias mitigation constraint, $L^\mathcal{A}$, we used squared distance between model's output and provided label instead of cross entropy loss since it is commonly used for these dataset. Cross entropy defined in equation~\ref{optim_loss} is more generic and provides similar results to squared distance. We chose baseline values $x'$ carefully, $-1$ and $-2$ to indicate missingness of the attributes in the dataset.
We compared our method with the well known GerryFair~\citep{pmlr-v80-kearns18a} and a mutual information reduction-based approach~\citep{fairness_mi} adopted for intersectional fairness. GerryFair performs a zero-sum optimization between a fairness auditor and a classifier subject to the auditor's constraints. On the other hand,~\citep{fairness_mi, song2019learning, Louppe2017LearningTP} show that mutual information reduction-based approaches can be formulated as generative adversarial optimization problems. Here the classifier plays the role of the generator and the discriminator aims to reduce the mutual information between the protected attributes and the output of the classifier. Inspired by~\citep{song2019learning, Louppe2017LearningTP} we implemented an adversarial network with two 2-layer RELU-BatchNorm-Linear NNs. One of those networks serves as a generator and the other one as a discriminator. 

\vspace{-1pt}
\subsection{Evaluation metrics}
\label{sec:eval_metrics}
As fairness measurement metrics we adopted Equalized Odds (EO)~\citep{Hardt2016EqualityOO, data_decisions} and Demographic Parity (DP)~\citep{dp2, dp3} metrics to measure model's intersectional subgroup biases.
Inspired by~\citep{data_decisions} and ~\citep{ghosh2021characterizing} we measure EO as the differences between minimum and maximum TPR and FPR scores across all subgroups formed by a given subset of protected attributes. Let $G$ denote the  set of different combinatorial options formed by the subgroups of protected attributes in the set $A_k \in \mathcal{A}$. For example $G = \{(Male, White), (Male, Black), (Male, Asian), (Female, White), (Female, Black), \\ (Female, Asian), ...  \}$. The values in $G$ are then denoted by $G_i \in G$. EO based on TPR and FPR are then defined as  $EO_G^{TPR} = | max(TPR(G_i)) - min(TPR(G_j)) |$ and $EO_G^{FPR} = | max(FPR(G_i)) - min(FPR(G_j)) |, G_i \in G, G_j \in G$ accordingly.
Similarly, demographic parity is measured as $DP_G = | max(DP(G_i)) - min(DP(G_j))|, G_i \in G, G_j \in G$.

\subsection{Experimental Results}
\begin{figure}[h]
  \centering
  \includegraphics[width=\linewidth]{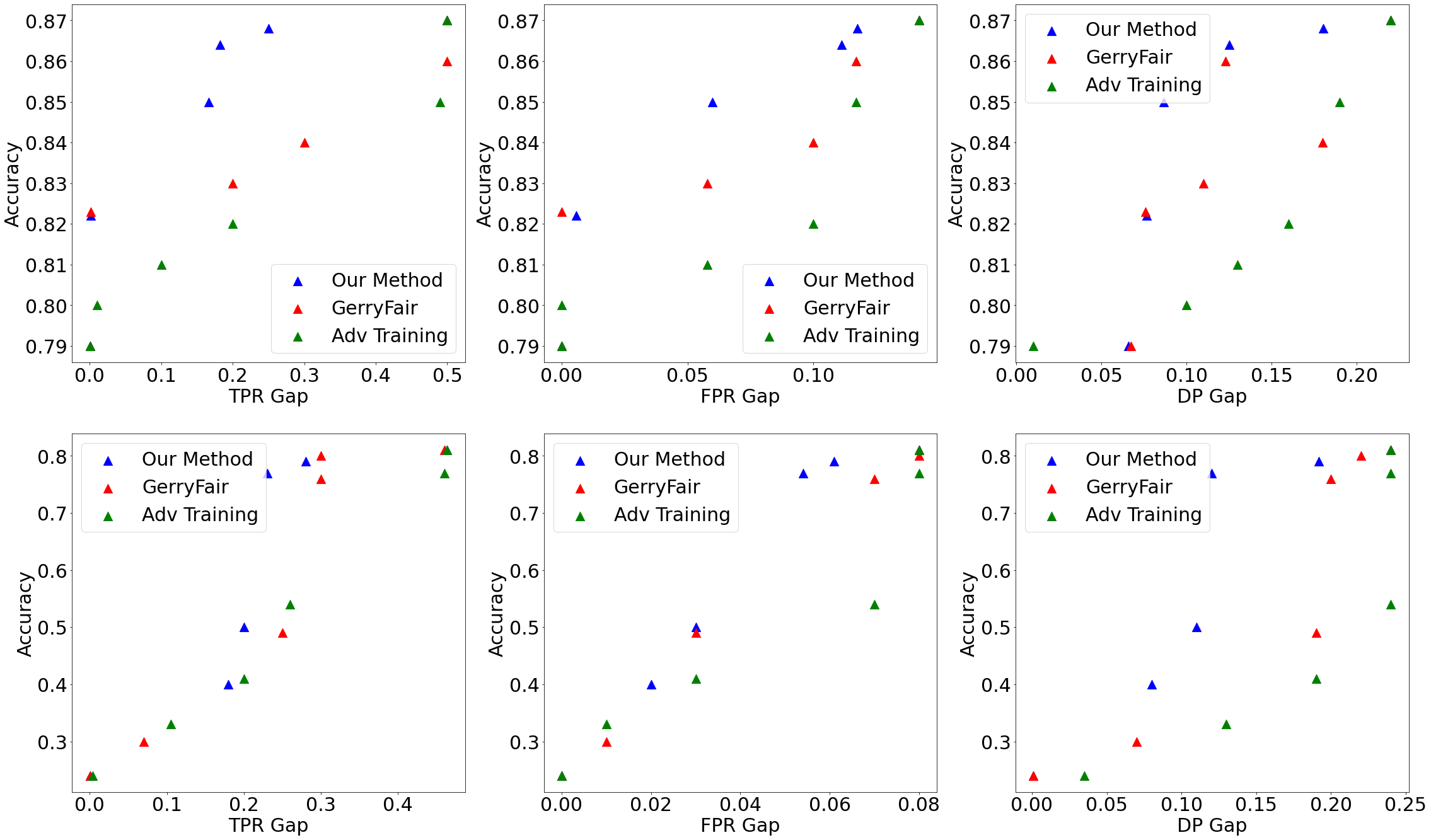}
  \caption{Our method, GerryFair and adversarially trained approaches compared on accuracy vs fairness metrics (TPR, FPR and DP gaps) for LSAC (top) and Adult (bottom) datasets. Best viewed in color.}
  \label{fig:our_vs_gerry_lsac}
\end{figure}

\begin{figure}
  \centering
  \includegraphics[width=\linewidth]{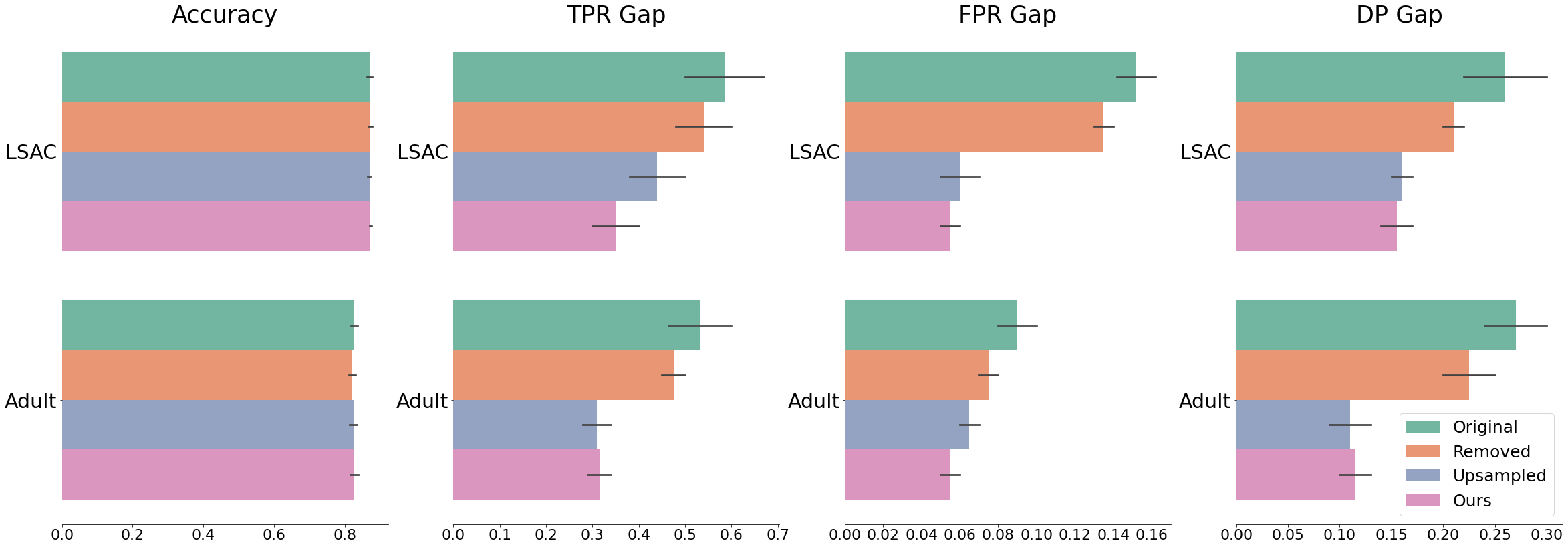}
  \caption{Our method ($\alpha=0.5$) vs original model, removed \texttt{Gender} and \texttt{Race} attributes and upsampled for underrepresented subgroups. Best viewed in color.}
  \label{fig:original_removed_upsamples_ours}
\end{figure}

In order to better understand the accuracy vs fairness gap relationship, we run multiple experiments by varying the weight of the fairness component from zero to a large number both for our approach and two baseline approaches. We compare our approach against GerryFair~\citep{pmlr-v80-kearns18a} and a mutual information-based approach~\citep{fairness_mi} in terms of Accuracy vs TPR, FPR and DP gaps. As described in section~\ref{sec:setup} mutual information based fairness mitigation is performed based on adversarial training. Our experimental results depicted on figure~\ref{fig:our_vs_gerry_lsac} reveal that our method results in higher accuracy when reducing TPR gap from 0.5 to 0.2 for LSAC and from 0.46 to 0.23 for Adult datasets. 
All three methods exhibit similar accuracy when the TPR gap is further reduced from approximately 0.2 to 0.0. We examine similar patterns for Accuracy vs FPR and DP gap measurements as well.
We observe that the adversarially trained approach slightly underperforms two other approaches in terms of accuracy. We hypothesize that generator-discriminator based approaches are more effective for binary protected attributes as also shown in~\citep{Louppe2017LearningTP}. When dealing with multiple non-binary protected attributes, generator-discriminator based networks become less effective and might require further fine tuning.
We conclude that our method can be beneficial especially when the goal is to reduce fairness gaps substantially without hurting the accuracy too much. Appendix~\ref{additional_experiments} provides additional results for the COMPAS~\citep{compas_ds_cite} dataset.
Furthermore, we compare our method with the original and two additional baseline models. One of the baselines represents a model trained without protected attributes. In this case we removed \texttt{Gender} \underline{and} \texttt{Race} attributes from the dataset. The second baseline model is trained on a dataset where examples for all underrepresented intersectional subgroups of \texttt{Gender} \underline{and} \texttt{Race} are upsampled. Figure ~\ref{fig:original_removed_upsamples_ours} summarizes Accuracy, TPR, FPR and DP gaps across all baselines and our approach for LSAC and Adult datasets. We observe that our method is comparable or better in reducing the fairness gap for almost the same accuracy trade-off.

In order to validate the effectiveness of our approach we perform two additional studies. We aim to understand how inference time masking of protected attributes and the presence of attributes strongly correlated with protected attributes, change the accuracy, TPR, FPR and DP gaps.

\paragraph{Masking protected attributes:}
\vspace{-5pt}
We mask protected attributes in the test dataset and compare model accuracy and fairness metrics before and after masking. This validates the hypothesis that fairness gap changes will remain changed before and after protected attributes are masked in the test dataset, if our method is effective in mitigating bias and is causally fair~\citep{causal_fair}.
Table~\ref{tab:TPRs_FPRs_DPs_masked} showcases the high sensitivity of the original model's accuracy, TPR, FPR and DP gaps when protected attributes are masked. In contrast to the original model, our model is almost insensitive to the masking of protected attributes showing its robustness to the presence of protected attributes. This also shows that privacy-preserving inference is possible since protected attributes are not required to be present at inference time.
Appendix~\ref{additional_experiments} demonstrates the results of the same experiment for Adult dataset.
\begin{table}
\vspace{-2pt}
  \caption{Test Accuracy, TPR, FPR and DP Passed Bar gaps for the Original and Our Models ($\alpha=0.5$) with and without masking of protected attributes applied on LSAC dataset. }
  \label{tab:TPRs_FPRs_DPs_masked}
  \centering
  \begin{tabular}{lcccc}
    \toprule
   & Accuracy & TPR & FPR & DP passed \\ 
  
    \midrule
    Original model & 0.87 & 0.55 &	0.14 &	0.23 \\
    Original model w/masked $A^{(Gender,Race)}$ & 0.85 & 0.53 & 0.05 & 0.15 \\
    Our model & 0.86 & 0.28 & 0.06 & 0.11 \\
    Our model w/masked $A^{(Gender,Race)}$ & 0.86 & 0.28 & 0.06 & 0.10 \\
  \bottomrule
\end{tabular}
\end{table}
\begin{table}
\vspace{-10pt}
  \caption{Accuracy and TPR computed for the original model, a baseline model trained without $\texttt{Gender}$ and $\texttt{Race}$, and our approach. The dataset contains an additional attribute $\texttt{Race1}$ for which mitigation is intentionally not performed.
  }
  \label{tab:corr_race_race1}
  \centering
  \begin{tabular}{cccc}
    \toprule
    & \multirow{2}{*}{Accuracy} & \multicolumn{2}{c}{TPR} \\ 
    \cmidrule{3-4} 
     &  &  $\texttt{Male} \cap \texttt{White}$  & $\texttt{Female} \cap \texttt{Black}$ \\
    \midrule
    Original model & 0.87 & 0.99 & 0.76\\
    Removed protected attributes & 0.86 & 0.96 & 0.77 \\
    Our model ($\alpha=0.5$) & 0.86 & 0.96 & 0.90\\
  \bottomrule
\end{tabular}
\end{table}

\paragraph{Studying the effects of strongly correlated features with protected attributes:}
\vspace{-10pt}
In order to understand the effectiveness of our approach in the presence of features strongly correlated with protected attributes we use two race related features \texttt{Race1} and \texttt{Race}. \texttt{Race1} is a coarse-grained representation of the \texttt{Race} attribute. We apply bias mitigation only to fields \texttt{Gender} and \texttt{Race}. No mitigation for field \texttt{Race1} is carried out. Table~\ref{tab:corr_race_race1} shows that our approach is still effective in mitigating the bias in underrepresented groups such as $\texttt{Female} \cap \texttt{Black}$. It is not as effective as in the absence of \texttt{Race1} field~\ref{tab:TPRs_Male_Female}, however, it is more effective than removing \texttt{Gender} and \texttt{Race} from the dataset.

\begin{figure}[h]
  \centering
  \includegraphics[width=10.0cm,
height=7.1cm]{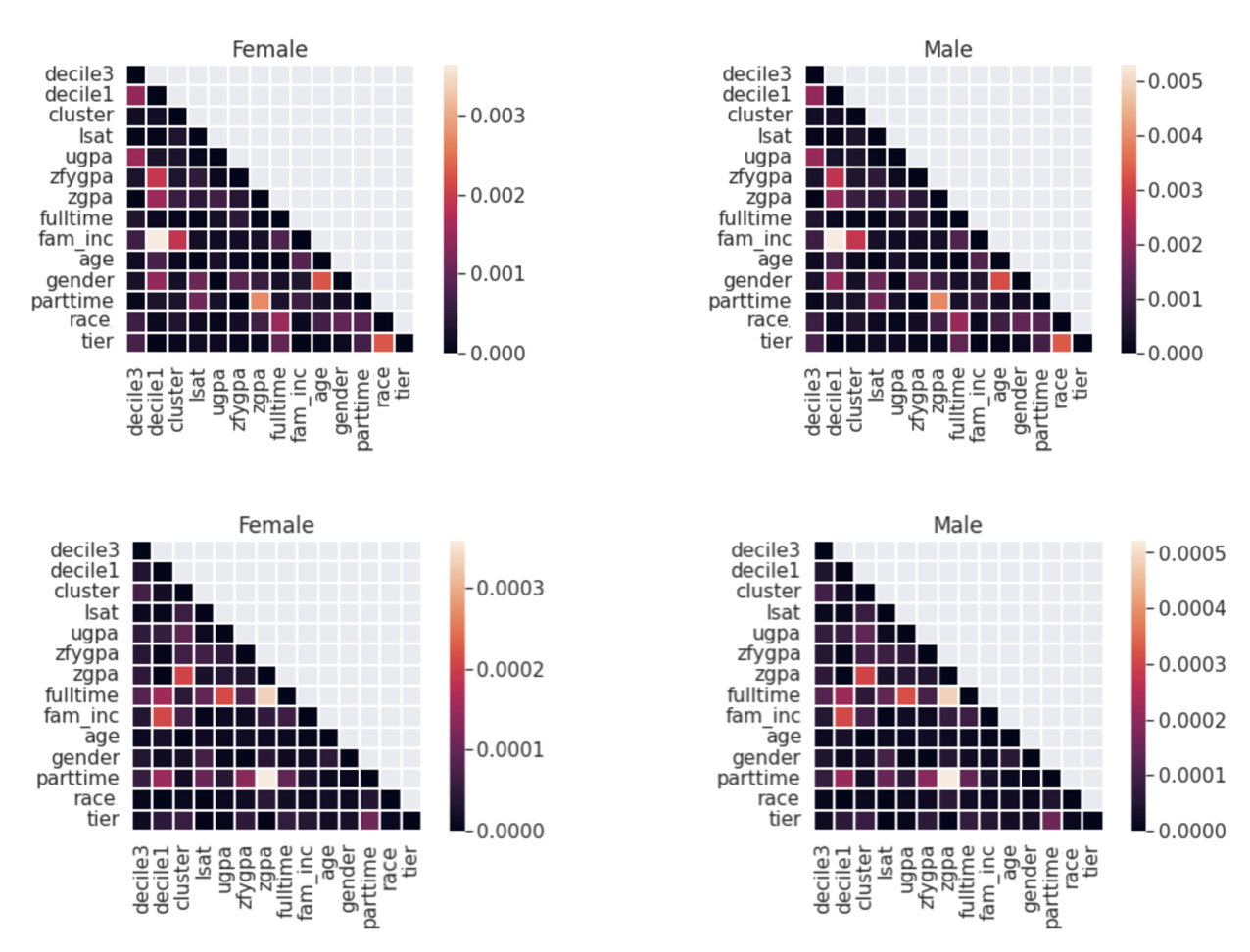}
  \caption{Aggregated pairwise feature interaction heatmaps before (top) and after (bottom) bias mitigation for both \texttt{Male} and \texttt{Female} based on LSAC test dataset. Best viewed in color.}
  \label{fig:finter_standard}
\end{figure}

\paragraph{Feature interaction effects:}
\vspace{-5pt}
In addition to accuracy, TPR, FPR and DP gaps, we also analyse pairwise feature interaction effects of protected attributes based on \cite{archattribute}. We validate the hypothesis that pairwise feature interaction of protected attributes with non-protected ones drops significantly for the unbiased model. Furthermore, our experiments reveal that the decline of feature interaction scores for protected attributes leads to the emergence of stronger interaction patterns between other attributes.
Figure~\ref{fig:finter_standard} visualizes aggregated pairwise feature interaction heatmaps for the original, biased model, at the top and unbiased model, based on our approach, at the bottom of the diagram. The results suggest that feature interaction effects for \texttt{Male} and \texttt{Female} look very similar. We also observe that the protected attribute \texttt{gender} has a relatively strong interaction pattern with the \texttt{age} attribute and \texttt{race} with the \texttt{fulltime} in the biased model. The unbiased model, however, exhibits no distinct and strong feature interaction patterns for \texttt{gender} and \texttt{race} with \texttt{age} and \texttt{fulltime} respectively. On the other hand, we discern stronger emerging interaction patterns between \texttt{parttime} and \texttt{zgpa}, \texttt{fulltime} and \texttt{zgpa}. This helps us better understand how feature interaction patterns are impacted when the models are trained with bias mitigation constraints. These findings can serve as sanity checks and facilitate better understanding of bias mitigation techniques.

\section{Conclusion and Future Work}
In this paper, we propose a framework to mitigate modeling bias in intersectional subgroups independent of the type of the protected attributes. The framework incorporates a generic proxy constraint into the optimization objective which reduces the mutual information between protected attributes and the output variable. We study accuracy vs TPR, FPR and DP gap trade-offs both with respect to state of the art approaches as well as data pre-processing techniques such as upsampling and removing protected attributes. We show empirically, that our approach surpasses GerryFair and adversarially trained approaches when reducing the fairness gaps.
Furthermore, we show, empirically, that our approach is still effective when other features are correlated with protected attributes and protected attributes are not available during inference time. Lastly, our experiments reveal that bias removal reduces the interaction effects between protected attributes and other attributes in the dataset.

In the future, we plan to analyse the effects of different types of bias mitigation constraints on a specific ML task of interest. This type of research can help better understand what kind of constraints might work better for a specific problem space and why. In addition to that it is also valuable to investigate techniques for estimating the optimal hyperparameter value for the weight of bias mitigation term, instead of a grid search approach.

\bibliographystyle{ACM-Reference-Format}
\bibliography{neurips_2022}

\appendix
\section{Appendix}

\subsection{Dataset}
\label{dataset_details}
In this section we describe LSAC and UCI Adult datasets in detail. Both datasets have a highly unbalanced distribution over their protected attributes such as race and gender.
Approximately $84\%$ of all samples in the LSAC dataset have \texttt{White} as race, while only 1.8\% have \texttt{Other} as race.
Likewise, there are more examples for \texttt{Male} than for \texttt{Female} gender. The labels have skewed distributions as well; with approximately 94\% samples labelled as \texttt{Passed} while only 6\% are labeled as \texttt{Not Passed}. The full breakdown of LSAC dataset is presented in Table~\ref{tab:lsac_ds}.
We observe similar data distribution patterns for the Adult dataset. Approximately $86\%$ of all samples in the Adult dataset are associated with \texttt{White} race, with only $0.7\%$ with $other$ race. There are more samples for \texttt{Male} than for \texttt{Female} and more sample for \texttt{<=50K} salary range than for \texttt{>50K}. The full breakdown of Adult dataset is presented in the table~\ref{tab:adult_ds}. Similar to LSAC and Adult dataset we observe uneven distribution of samples across  \texttt{Gender} and \texttt{Race} intersectional subgroups~\ref{tab:compas_ds}. We observe that, specifically, the number of examples with recidivism label for \texttt{Male} and \texttt{Not Caucasian} subgroup surpass the number of examples with no recidivism label for the same subgroup. This is not the case for any other subgroup formed by \texttt{Gender} and \texttt{Race} attributes.

\begin{table}[h]
  \caption{LSAC dataset breakdown by gender, race and dataset label (Bar Pass and Bar Not Pass)}
  \label{tab:lsac_ds}
  \begin{tabular}{c | c | c | c | c | c }
     & Black & Hispanic &	Asian &	White &	Other \\
    \midrule
    Male ( Not Pass / Pass )	& 99 / 352 &	64 / 443 &	33 / 363 &	311 / 9622	& 17 / 197 \\
    \midrule
    Female ( Not Pass / Pass ) &	167 / 580 &	51 / 368 &	27 / 367 &	249 / 6957 &	20 / 140 \\
  \bottomrule
\end{tabular}
\end{table}

\begin{table}[h]
  \caption{Adult dataset breakdown by gender, race and income category}
  \label{tab:adult_ds}
  \begin{tabular}{c | c | c | c | c | c }
     & Black & Asian-Pac-Isl & Amer-Ind-Esk & White & Other \\
    \midrule
    Male($<=$50k/$>$50k)	& 1736 / 408 &	563 / 304 &	230 / 39 &	18268 / 8752	& 191 / 36 \\
    \midrule
    Female($<=$50k/$>$50k) &	1958 / 126 &	371 / 65 &	152  / 14 &	10428 / 1455 &	117 / 9 \\
  \bottomrule
\end{tabular}
\end{table}

\begin{table}[hbt!]
  \caption{COMPAS dataset breakdown by gender, race and recidivism label}
  \label{tab:compas_ds}
  \centering
  \begin{tabular}{c | c | c }
     & Caucasian & Not Caucasian \\
    \midrule
    Male(No recid. / Did recid.)	& 968 / 652 &	1630 / 1744 \\
    \midrule
    Female(No recid. / Did recid.) &	310 / 170 &	450 / 230 \\
  \bottomrule
\end{tabular}
\end{table}

\subsection{Additional Experiments}
\label{additional_experiments}
Similar to LSAC dataset, we perform additional experiments on the Adult dataset when protected attributes \texttt{Gender} and \texttt{Race} are masked in the test dataset. We observe that after masking those attributes, Accuracy, TPR, FPR and DP Passed gaps do not change much. This validates the hypothesis that our method learns latent representations that do not rely on protected attributes.

\begin{table}[h]
  \caption{Test Accuracy, TPR, FPR and DP Passed Bar gaps for the original and our models ($\alpha=0.5$) with and without masking of protected attributes applied on Adult dataset. }
  \label{tab:TPRs_FPRs_DPs_masked_adult}
  \centering
  \begin{tabular}{lcccc}
    \toprule
   & Accuracy & TPR & FPR & DP Passed \\ 
  
    \midrule
    Original model & 0.81 & 0.46 &	0.08 &	0.24 \\
    Original model w/masked $A^{(Gender,Race)}$ & 0.80 & 0.40 & 0.06 & 0.19 \\
    Our model & 0.80 & 0.23 & 0.06 & 0.19 \\
    Our model w/masked $A^{(Gender,Race)}$ & 0.80 & 0.22 & 0.06 & 0.19 \\
  \bottomrule
\end{tabular}
\end{table}

\begin{figure}[h]
  \centering
  \includegraphics[width=\linewidth]{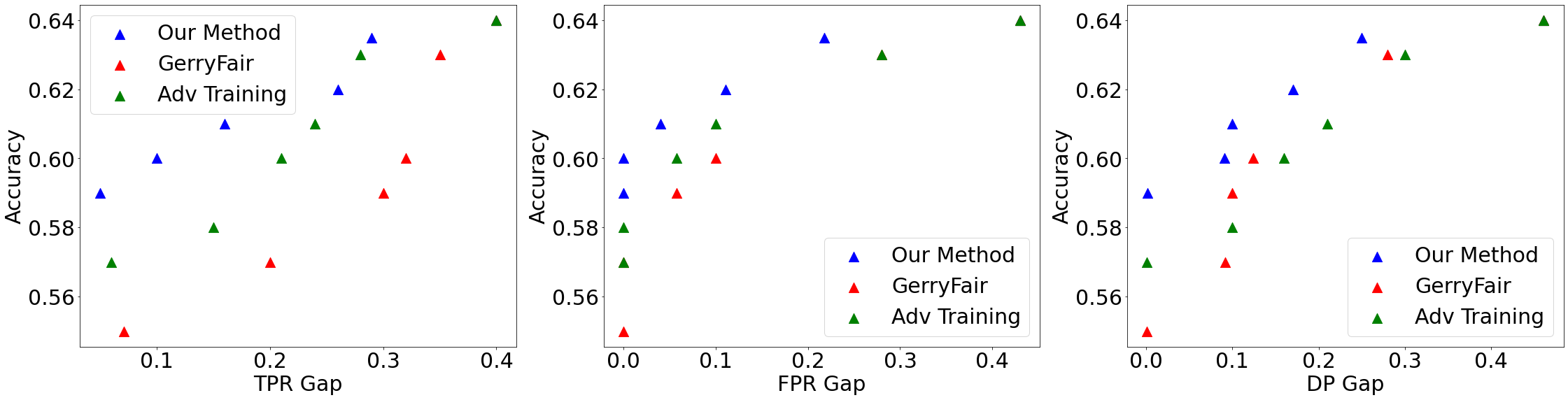}
  \caption{Our method, GerryFair and adversarially trained approaches compared on accuracy vs fairness metrics (TPR, FPR and DP gaps) for COMPAS dataset. Best viewed in color.}
  \label{fig:our_vs_gerry_compas}
\end{figure}

Similar to LSAC and Adult datasets, we perform additional experiments with COMPAS dataset in order to further support our findings and empirical evidence. Figure~\ref{fig:our_vs_gerry_compas} compares our method against GerryFair and adversarially-trained methods when mitigating the intersectional subgroup bias for \texttt{Gender} and \texttt{Race} protected attributes.
We observe that our approach outperforms GerryFair and adversarially trained methods in terms of reducing Accuracy vs TPR, FPR and DP gap tradeoffs.

\end{document}